\documentclass[10pt,twocolumn,letterpaper]{article}
\usepackage[T1]{fontenc}
\usepackage{CJKutf8}
\usepackage[pagenumbers]{cvpr}
\usepackage{array,tabularx}
\usepackage{caption}
\definecolor{cvprblue}{rgb}{0.08,0.20,0.36}
\usepackage[pagebackref,breaklinks,colorlinks,hyperfootnotes=false,allcolors=cvprblue]{hyperref}
\def\confName{CVPR}
\def\confYear{2027}
\hypersetup{pdftitle={Contact Is Not Capture: Benchmarking and Learning Stable Acquisition of Free-Floating Objects in Space},pdfauthor={Zizhe Wang, Jianing Li, Yao Zhu, Yunjian Zhang, Mengyi Yan, Jingzheng Li, Yi Huang, Jia Deng}}
\title{Contact Is Not Capture: Benchmarking and Learning Stable Acquisition of Free-Floating Objects in Space}
\author{Zizhe Wang$^{1}$\quad Jianing Li$^{2}$\quad Yao Zhu$^{3}$\quad Yunjian Zhang$^{4}$\\
Mengyi Yan$^{5}$\quad Jingzheng Li$^{6}$\quad Yi Huang$^{3}$\quad Jia Deng$^{1}$\\[2pt]
{\normalsize $^{1}$Astroup Lab\quad $^{2}$Peking University\quad $^{3}$Harbin Institute of Technology}\\
{\normalsize $^{4}$University of Chinese Academy of Sciences\quad $^{5}$Shandong University\quad $^{6}$Zhongguancun Laboratory}\\[2pt]
{\tt\footnotesize \{wangzizhe, dengjia\}@astroup.com\quad lijianing@pku.edu.cn\quad ee\_zhuy@zju.edu.cn}\\
{\tt\footnotesize zhangyunjian@iie.ac.cn\quad yanmy@sdu.edu.cn\quad lijingzheng@zgclab.edu.cn\quad 24B903024@stu.hit.edu.cn}}
\begin{document}
\raggedbottom
\setcounter{dbltopnumber}{2}
\renewcommand{\dbltopfraction}{0.88}
\setlength{\dbltextfloatsep}{12pt plus 2pt minus 2pt}
\setlength{\textfloatsep}{12pt plus 2pt minus 2pt}
\setlength{\dblfloatsep}{10pt plus 2pt minus 2pt}

\twocolumn[{
\renewcommand\twocolumn[1][]{#1}
\vspace*{-16pt}
\maketitle
\vspace{-24pt}
{\centering
  \includegraphics[width=0.91\textwidth]{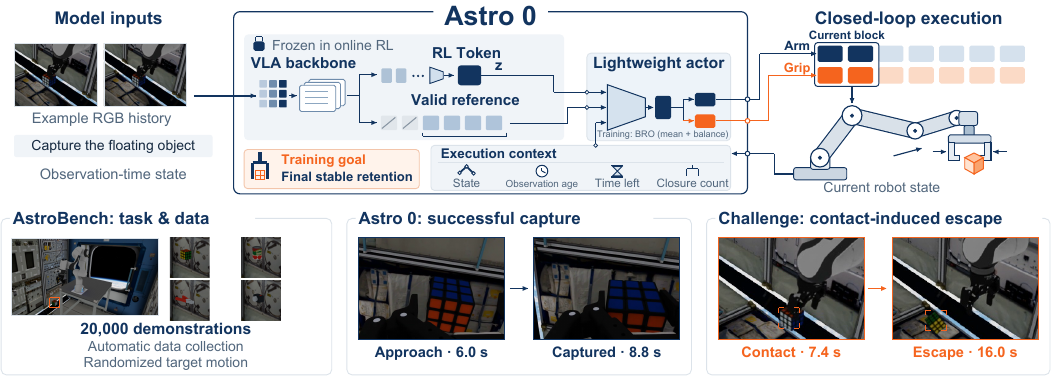}\par
  \vspace{2pt}
  \captionsetup{hypcap=false}
  \captionof{figure}{\textbf{Astro 0 and AstroBench.} A frozen VLA supplies features and reference actions to an ECC-conditioned actor trained with BRO. Below: benchmark resources, successful capture, and contact-induced escape in separate episodes. Times are elapsed within each sequence.}
  \label{fig:astrobench}
  \captionsetup{hypcap=true}
\par}
\vspace{6pt}
}]
{\renewcommand{\thefootnote}{}\footnotetext{$^{1}$Astroup Lab: \begin{CJK*}{UTF8}{gbsn}向上跃升（北京）科技有限公司\end{CJK*} (\href{https://astroup.com}{\texttt{astroup.com}}).}}

\begin{abstract}
  Capturing free-floating objects is a basic capability for sample transfer and equipment servicing in space. Unlike supported objects on Earth, floating targets can be pushed out of reach before a stable grasp forms, and their motion continues during perception and inference. We introduce AstroBench, a simulated benchmark in which a fixed-base arm inside a spacecraft cabin captures free-floating objects in zero gravity. Simulation runs continuously during inference, and evaluation separates contact, stable retention, and target loss: learned models contact 95.3--97.7\% of targets but stably retain only 80.1--87.8\%. An automated pipeline provides 20,000 successful demonstrations across 20 object groups; we show that success filtering underrepresents distant targets and correct this with distance-stratified sampling. To handle stale observations and poor approach--closure coordination, we develop Astro 0, a lightweight actor on a frozen VLA that conditions on observation age and current robot state (ECC) and optimizes mean actions with balanced arm and gripper reference constraints (BRO). On the standard dynamic capture task, Astro 0 reaches 87.8\% stable retention, 4.61 percentage points above Direct RLT, our DynamicVLA + RL Token baseline, under matched training and execution conditions.
\end{abstract}

\section{Introduction}
\label{sec:introduction}
Space robots must capture and secure objects for sample handling and equipment servicing when continuous human supervision is difficult~\cite{spacefm2026}. Success requires grasping and retaining a freely moving target, not merely touching it.

Free-floating capture in space differs from supported grasping on Earth in two coupled ways. Targets move throughout observation, inference, and execution, making observations stale. Without support to resist incidental contact, approaching fingers can push a target away before closing: contact does not imply acquisition. In a direct-transfer pilot without task-specific fine-tuning, $\pi0.5$ succeeded in only 2 of 100 trials. A single success rate hides these distinct failures.

Existing resources do not measure this contact-to-retention transition. GraspNet and RoboTwin evaluate supported objects~\cite{graspnet2020,robotwin2025}, and Dynamic Object Manipulation (DOM) studies moving targets~\cite{dynamicvla2026} but does not record whether contact leads to stable retention in zero gravity. Success-filtered demonstration pipelines rarely examine the resulting bias. DynamicVLA~\cite{dynamicvla2026} and RL Token~\cite{rlt2026} address inference latency and online refinement, but not how the actor should account for observation age or weight the gripper against the arm. Free-floating capture thus requires changes in evaluation, collection, and action refinement.

AstroBench evaluates a fixed-base arm inside a spacecraft cabin capturing unsupported targets in zero gravity, with simulation continuing during inference. Its two-layer protocol distinguishes contact, final stable retention (the task endpoint), and irreversible workspace exits, including exits before contact. First-closure retention (FCR) measures success with one effective closure. The distinction matters: learned models contact 95.3--97.7\% of targets but stably retain only 80.1--87.8\%.

Our pipeline provides 20,000 successful demonstrations across 20 object groups in LeRobot format, with explicit splits and acceptance records. Success filtering underrepresents distant targets; distance-stratified sampling corrects this bias. Equal-size studies examine motion coverage across three model families.

Astro 0 combines DynamicVLA's asynchronous execution~\cite{dynamicvla2026} with RL Token's compressed interface~\cite{rlt2026}. A lightweight actor on a frozen VLA refines actions every 80 ms. Execution-Context Conditioning (ECC) supplies observation age, remaining time, and closure history. Balanced Reference-Guided Optimization (BRO) optimizes mean actions with equal arm and gripper reference weights. Training rewards final stable retention, allowing retries while the target remains reachable.

Astro 0 reaches 87.8\% stable retention, 4.61 points above Direct RLT (DynamicVLA + RL Token) under matched conditions. Component controls separate ECC and BRO, while visual-delay and transfer tests examine stale observations and generalization.

We make three contributions:

\noindent\textbf{(1) A benchmark for continuous free-floating acquisition.} AstroBench separates contact, stable retention, and all-cause target loss, exposing an acquisition gap that persists across contact-physics settings.

\noindent\textbf{(2) A capture dataset with collection-bias correction.} We provide 20,000 demonstrations with acceptance records, identify distance-dependent selection bias, and correct it through stratified sampling.

\noindent\textbf{(3) Execution-aware capture refinement.} Astro 0 introduces ECC and BRO to coordinate approach and closure under stale observations. Matched controls test both components, and held-out motions and geometries test transfer.

\section{Related Work}
\label{sec:related}
\paragraph{Space manipulation and evaluation resources.}
Space reviews discuss foundation models and autonomous science~\cite{spacefm2026,spacescience2026}; predictive and two-phase visual servoing address moving targets~\cite{dong2018,zhang2024}. GraspNet labels and evaluates cluttered grasps~\cite{graspnet2020}, RoboTwin synthesizes dual-arm demonstrations~\cite{robotwin2025}, and DOM collects moving-object demonstrations and evaluates dynamic interaction~\cite{dynamicvla2026}. AstroBench instead records whether contact with an unsupported target becomes stable retention, alongside collection acceptance records (Table~\ref{tab:resource-comparison}).

\paragraph{Generalist models and dynamic grasping.}
OpenVLA and $\pi0.5$ transfer visuomotor skills across tasks~\cite{openvla2024,pi05}. ACT and Diffusion Policy predict action sequences~\cite{act2023,diffusionpolicy2023}; SmolVLA combines a compact model with asynchronous execution~\cite{smolvla2025}. Reactive grasping maintains target-referenced correspondences~\cite{reactive2023}, while GraspLDP uses grasp-pose and graspness priors~\cite{graspldp2026}. AstroBench evaluates direct transfer and task adaptation separately, measuring whether the target remains held after contact.

\paragraph{Delay-aware execution and online refinement.}
Real-time chunking preserves committed actions during asynchronous inference~\cite{rtc2025}; DynamicVLA's Continuous Inference (CI) and Latency-aware Action Streaming (LAAS) remove expired action prefixes~\cite{dynamicvla2026}. These schedules compensate for delay without explicitly conditioning the actor on observation age. Residual RL learns control corrections~\cite{residualrl2018}, and RL Token compresses VLA features for reference-conditioned actor--critic learning~\cite{rlt2026}. Astro 0 retains these execution and representation interfaces. ECC supplies execution context; BRO combines deterministic mean-action updates~\cite{td32018} with balanced arm--gripper reference errors. Unlike FAN's output-distribution regularization~\cite{fan2026}, BRO constrains action groups against aligned references.

\section{AstroBench: Task, Evaluation, and Data}
\label{sec:benchmark}
\begin{table}[!t]
\centering\small
\caption{Representative resources and their evaluation focus. Entries summarize the cited versions; they are not an exhaustive feature checklist.}
\label{tab:resource-comparison}
\setlength{\tabcolsep}{3pt}
\begin{tabularx}{\columnwidth}{@{}p{0.23\columnwidth}p{0.23\columnwidth}X@{}}
\toprule Resource & Target setting & Learning / evaluation focus\\\midrule
GraspNet~\cite{graspnet2020} & Cluttered objects & Grasp-pose labels; analytic grasp evaluation\\
RoboTwin~\cite{robotwin2025} & Dual-arm manipulation & Digital-twin demonstrations; task success\\
DOM~\cite{dynamicvla2026} & Moving objects & Automatic demonstrations; dynamic interaction and transfer\\
\textbf{AstroBench} & \textbf{Unsupported, zero gravity} & \textbf{Capture demonstrations; contact, retention, and exit accounting}\\
\bottomrule
\end{tabularx}
\end{table}

AstroBench brings together a capture task, an evaluation protocol, a demonstration dataset, and a set of evaluation suites. Section~3.1 defines free-floating capture, and Section~3.2 separates episode outcomes from contact and target-loss events. Section~3.3 describes automatic collection, data splits, and acceptance records, while Section~3.4 defines the standard, diagnostic, and generalization evaluations.

\begin{figure*}[!t]
\centering
\includegraphics[width=\textwidth]{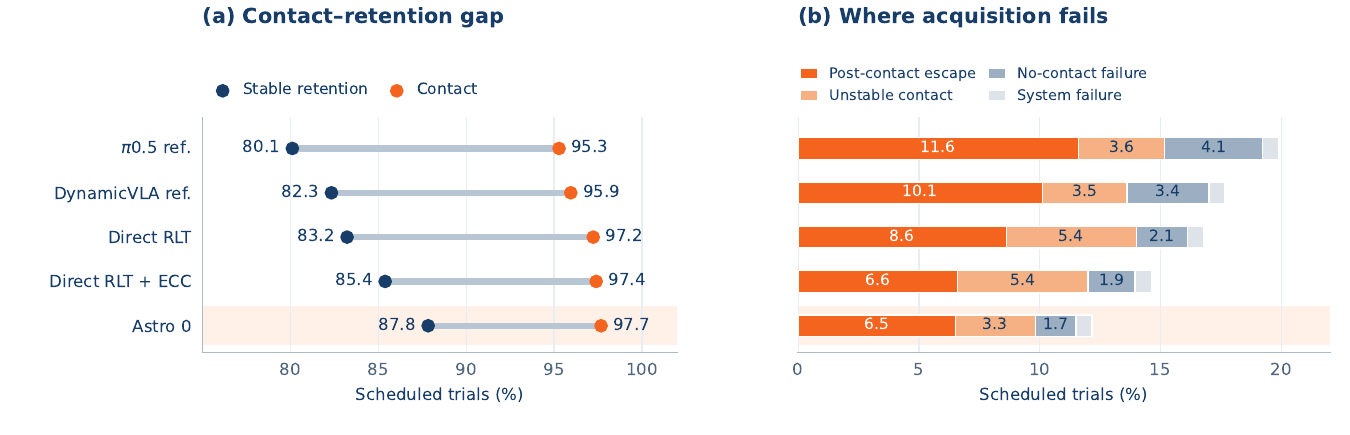}
\caption{\textbf{Contact is not acquisition.} Task-adapted models on 1,800 standard trials each. (a) Contact coverage and stable retention. (b) Mutually exclusive failure outcomes, as percentages of all scheduled trials; segments sum to $100\%$ minus retention.}
\label{fig:task-gap}
\end{figure*}

\subsection{Free-floating capture}
Each episode contains a fixed-base arm, a gripper, and a rigid target under zero gravity. The adapter uses Isaac Sim 4.5.0 with Isaac Lab 2.0.0 and the PhysX TGS rigid-contact solver, with eight position and two velocity iterations. The arm starts from a fixed configuration, while target position and initial linear and angular velocity vary. Subsequent motion follows rigid-body and contact dynamics. The deployed model receives external and wrist RGB, robot proprioception, and a target-specific instruction; the collector and scorer additionally access simulator object state.

The protocol uses a 12-s horizon. The adapter runs physics at 200 Hz, commands at 25 Hz, and an independent scoring trace at 50 Hz. Simulation advances during computation and communication. The four standard tracks are floating static, drift at 0.02 m/s, translation at 0.02 m/s with 0.5 rad/s rotation, and fast tumbling at 0.05 m/s with 1.0 rad/s rotation. The dynamic suite pools the last three tracks. In the nominal tracks, targets translate along the signed world-X axis and rotate about the signed world-Z axis; configurable extensions support arbitrary three-dimensional directions.

The evaluated subset contains 20 core object groups and 20 additional instances. The broader asset inventory also includes MELFI V6/RRM TB1 storage and servicing workcells, extending the catalog beyond free-floating capture.

\subsection{Retention, contact, and target loss}
\paragraph{Two-layer outcome accounting.}
The first layer assigns each episode one mutually exclusive outcome: retention, post-contact escape, unstable contact, no-contact failure, or system failure. The second layer independently records valid contact $C$ and irreversible workspace exit $L$, including events in episodes labeled as system failures. Valid contact requires at least one finger to exert 0.5 N for 10 ms, confirmed strictly before any exit. Thus, $\Pr(L)$ counts exits before contact as well as exits during system failures. All scheduled trials remain in the primary denominator. Physical logging continues through recoverable controller or API errors, and missing traces are reported separately.

\paragraph{Stable retention.}
The primary endpoint, $S_{\mathrm{retain}}$, requires measured closure and bilateral contact of at least 0.5 N per finger throughout the final second, with no forbidden contact or irreversible workspace exit. Relative linear and angular speeds must remain at most 0.02 m/s and 0.2 rad/s, respectively, while translation and rotation excursions must stay within 0.01 m and 0.1 rad.

\paragraph{First-closure retention.}
As an auxiliary metric, FCR measures whether the target is stably retained with one effective closure:
\begin{equation}
S_{\mathrm{first}}(\mathcal T)=S_{\mathrm{retain}}(\mathcal T)\land[N_{\mathrm{close}}(\mathcal T)=1].
\label{eq:success}
\end{equation}
An effective closure requires 2 mm of measured closing travel; 3 mm of reopening rearms the counter. Reopening and reclosing constitutes a second attempt, whereas multiple physical contacts do not. A second closure sets FCR to zero but allows the episode to continue. Stable retention measures eventual acquisition regardless of closure count.

A model may produce fewer post-contact escapes simply because it contacts fewer targets. We therefore report contact coverage $\Pr(C)$, retention conditional on contact $\Pr(S_{\mathrm{retain}}\mid C)$, and all-cause target loss $\Pr(L)$. Together, these measures distinguish successful acquisition from avoiding contact or losing a target before reaching it.

\begin{figure*}[!t]
\centering
\includegraphics[width=\textwidth]{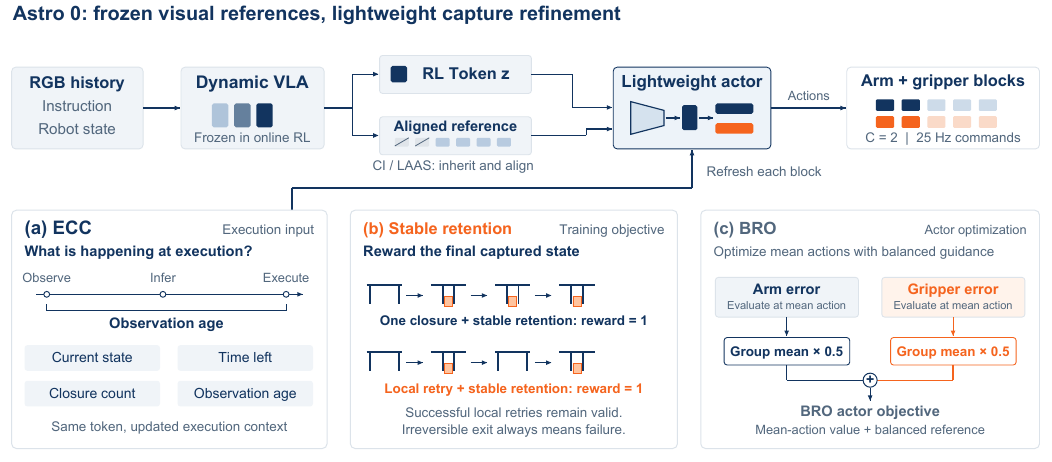}
\caption{\textbf{Astro 0: execution-conditioned learning for stable acquisition.} A frozen VLA backbone supplies a visual token and time-aligned reference actions to the actor. (a) ECC refreshes execution-time state and timing information. (b) The terminal reward accepts any final stable grasp, including a local retry before escape. (c) BRO optimizes mean actions using critic values and equally weighted arm and gripper reference errors. The reward and optimization objective are used only during training.}
\label{fig:architecture}
\end{figure*}

\subsection{Automatic collection and learning data}
At 25 Hz, the collector predicts 32 aperture-compatible antipodal grasps from simulator geometry and motion, searching intercepts from 80 ms to 2 s on a 40-ms grid. It selects the earliest collision-free, reachable candidate. Within the 30-mm approach corridor, it keeps the selected object-relative grasp fixed and closes only when the predicted enclosure margin reaches 4 mm. Tracking continues through closure, allowing two local retries while preserving the simulated dynamics. We retain 20,000 successful demonstrations from 32,000 collection attempts, an acceptance rate of 62.5\%, with 1,000 trajectories per group balanced across four tracks. Stored RGB, timestamps, joint states, commands, object annotations, and acceptance metadata follow explicit coordinate and timing conventions.

Each group contributes 800/100/100 trajectories to training, validation, and offline testing, giving 16,000/2,000/2,000 trajectories overall. Whole trajectories and related or near-duplicate episodes stay in the same partition, and normalization uses training data only. The standard split shares object groups across partitions; separate holdout tests exclude geometry groups.

\paragraph{Acceptance audit.}
We retain failed draws in the collection records to examine which initial states are underrepresented among successful demonstrations. The outer quartile of initial target distance accounts for 25\% of attempted draws but only 19\% of accepted demonstrations. Finite search, fixed grasp faces during approach, and closure gating limit the collector even with simulator-state inputs. Closed-loop evaluation samples fresh states from the original reset distribution without filtering them by collector success.

\subsection{Evaluation suites}
\label{sec:benchmark-transfer}
\paragraph{Standard dynamic capture.}
Three dynamic tracks provide 600 initial states per seed; three seeds yield 1,800 executions per model for Section~5's main comparison.

\paragraph{Diagnostic suites.}
Direct transfer uses 100 pilot trials per model without task-specific fine-tuning (Section~5.1). Contact-physics tests vary one factor at a time over 600 matched states per seed (Section~5.1). Visual-delay tests use 200 states per delay--rotation setting (Section~5.3). Both controlled suites use three seeds.

\paragraph{Generalization suites.}
Models trained on 4,000 demonstrations face two withheld motion combinations, with 400 states per seed. Frozen standard models face 20 new geometries, with 600 states per seed. Both use three seeds (Section~5.3); Table~S1 summarizes training support, sizes, and shared controls.

\section{Astro 0: Capture-Oriented Learning}
\label{sec:method}
DynamicVLA produces visual features and a sequence of reference actions using CI/LAAS~\cite{dynamicvla2026}. Astro 0 uses RL Token's compressed representation~\cite{rlt2026} and a lightweight actor to update the next two commands every 80 ms. Execution context accounts for observation age, while BRO combines mean-action optimization with balanced arm and gripper reference constraints. The training objective is final stable retention (Fig.~\ref{fig:architecture}).

\subsection{Dynamic proposals and execution context}
The backbone proposes a reference action sequence and visual token $z_{j(t)}$. LAAS removes expired reference rows. The actor receives
\begin{equation}
\begin{aligned}
x_t&=(z_{j(t)},p_t,\bar A_t,\Delta_t,\tau_t,b_t),\\
A_t&=\mu_\theta(x_t)\in\mathbb R^{C\times d_a},\quad C=2,
\end{aligned}
\label{eq:actorinput}
\end{equation}
where $p_t$ is current proprioception and $\bar A_t$ the aligned reference. ECC adds observation age $\Delta_t$, normalized remaining time $\tau_t$, and clipped effective-closure count $b_t=\min(N_{\mathrm{close}}(t),2)$. Actor and critic share this context. Object pose and velocity are excluded.

Observation age is acquisition-to-scheduled-execution time, $\Delta_t=t_{\mathrm{exec}}^{\mathrm{scheduled}}-t_{\mathrm{obs}}$, for the latest RGB frame in the proposal. Remaining time tells the actor how long it has to complete the task; the closure count records previous attempts. Current state and context refresh at each 80-ms block boundary, even when the visual token is unchanged. Each block contains two 25-Hz commands and finishes before switching proposals. Matched RLT controls share this schedule. Cameras continue sampling, while the actor reuses the latest visual token until the backbone publishes a new one.

\subsection{Learning final stable acquisition}
The model learns to establish a stable grasp by the task deadline:
\begin{equation}
\begin{aligned}
r_t&=\mathbb I[t=N-1]S_{\mathrm{retain}}(\mathcal T),\quad N=300,\\
J_{\mathrm{retain}}(\theta)&=\mathbb E_{\mathcal T\sim\mu_\theta}[S_{\mathrm{retain}}(\mathcal T)].
\end{aligned}
\label{eq:reward}
\end{equation}
Here $t$ indexes 25-Hz steps and each actor decision covers two. The retention predicate is defined in Section~\ref{sec:benchmark}. Reopening and reclosing may succeed while the target remains reachable; crossing the escape boundary is an irreversible failure. A second closure neither terminates the episode nor zeros a later retention reward.

All online comparators use this reward, the same 12-s horizon, and identical physical failure rules, without distance, contact, or speed shaping. FCR remains a secondary diagnostic and an optional reward control. Closure history supplies interaction context without imposing a single-attempt constraint.

\begin{table*}[!t]\centering\small
\caption{Standard dynamic capture under common two-step execution. Rates are percentages; brackets give 95\% seed-$t$ intervals across three training seeds (1,800 trials/model). All metrics describe the same episodes. RGB predictive capture additionally uses known geometry and pose-label supervision. Online models share the final-retention reward. Direct RLT is the base comparator; Direct RLT + ECC is the execution-context component control.}
\label{tab:capture-overview}
\setlength{\tabcolsep}{3.5pt}
\begin{tabular}{lrrrrr}\toprule
Model & Retention & FCR & Target loss & Contact & Ret./contact\\\midrule
RGB predictive capture & 72.9 [70.5,75.3] & 69.9 [67.5,72.3] & 15.4 [13.0,17.9] & 93.9 & 77.6\\
$\pi0.5$ ref. & 80.1 [77.1,83.2] & 75.1 [71.7,78.6] & 13.9 [12.3,15.5] & 95.3 & 84.1\\
DynamicVLA ref. & 82.3 [80.1,84.5] & 79.1 [76.7,81.4] & 12.2 [10.4,14.1] & 95.9 & 85.8\\
Direct RLT & 83.2 [80.0,86.4] & 76.1 [72.5,79.6] & 10.3 [8.1,12.5] & 97.2 & 85.6\\
Direct RLT + ECC & 85.4 [82.9,87.9] & 79.0 [75.7,82.3] & 8.4 [6.7,10.2] & 97.4 & 87.7\\
\textbf{Astro 0} & 87.8 [85.9,89.7] & 83.2 [80.7,85.7] & 7.7 [6.2,9.2] & 97.7 & 89.9\\
\bottomrule\end{tabular}
\end{table*}

\subsection{Balanced reference-guided optimization}
Demonstration adaptation precedes token compression and online actor--critic learning. The adapted backbone and Token encoder remain frozen during online learning. Balanced Reference-Guided Optimization (BRO) combines two choices in one actor objective: optimizing the mean action used at deployment and balancing arm and gripper reference errors.

Let $A_\theta=\mu_\theta(x_t^m)$ denote the actor's mean action block, where $x_t^m$ uses reference dropout during training. The critic receives the unmasked context $x_t$, and the aligned reference $\bar A_t$ remains the target. We optimize
\begin{equation}
\begin{aligned}
\mathcal L_{\mathrm{BRO}}&=\mathbb E_{\mathcal B}\bigl[-Q_{\psi_1}(x_t,A_\theta)\\
&\qquad+\lambda_{\mathrm{ref}}\mathcal L_{\mathrm{bal}}(A_\theta,\bar A_t)\bigr],\\
\mathcal L_{\mathrm{bal}}&=\tfrac12\operatorname{MSE}_{\mathrm{arm}}(A_\theta,\bar A_t)\\
&\qquad+\tfrac12\operatorname{MSE}_{\mathrm{grip}}(A_\theta,\bar A_t).
\end{aligned}
\label{eq:reference}
\end{equation}
Each reference error averages normalized coordinates within its action group and valid block steps. Uniform averaging over eight coordinates assigns only one eighth of the reference weight to the gripper; balanced averaging assigns half. Both the value term and the reference penalty are evaluated at the mean action, following deterministic actor--critic optimization~\cite{td32018}. This aligns training with deployment while retaining arm and gripper guidance. Collection adds exploration noise; evaluation uses the mean action. We cross mean and stochastic updates with balanced and uniform penalties to separate the two components of BRO under equal reference-weight tuning budgets.

\section{Experiments}
\label{sec:experiments}
\paragraph{Shared setup.}
Training uses eight A100 GPUs, 16,000 demonstrations, and 120k control steps per seed for online models. Seeds 17/29/43 each evaluate 20 groups $\times$ ten starts $\times$ three tracks (1,800 executions/model). Methods share resets, validation-retention selection, and a time-aligned two-step interface. Simulation continues during inference. We report means and descriptive 95\% seed-$t$ intervals; paired differences match seeds and states.

The $\pi0.5$ and DynamicVLA references use supervised adaptation. The RGB predictive baseline estimates pose and velocity, extrapolates motion, and tracks geometry-derived grasps; it uses known geometry and pose-label supervision, but no deployed target-state ground truth. Direct RLT combines DynamicVLA, CI/LAAS, and RL Token. Direct RLT + ECC adds context; Astro 0 also adds BRO. All three share the checkpoint, actor capacity, initialization sources, final-retention reward, and online budget. Reference weights are tuned separately with equal validation budgets. Main results use natural demonstration sampling.

\subsection{Contact is not stable acquisition}
\paragraph{Direct transfer.}
Without task-specific fine-tuning, $\pi0.5$ succeeds in 2/100 pilot trials and GPT6-Astra~\cite{astranomy} in 0/100. Its long inference time allows targets to leave before effective action. After task adaptation, Fig.~\ref{fig:task-gap} separates the contact--retention gap from the failure outcomes.

Learned models contact 95.3--97.7\% of targets but retain only 80.1--87.8\% (Table~\ref{tab:capture-overview}). The RGB predictive pipeline retains 72.9\%. Astro 0 gains 4.61 points over Direct RLT (83 additional retentions). Its 138 target losses comprise 117 post-contact, 15 uncontacted, and six technical-failure exits. Retention conditional on contact is 89.9\%, versus DynamicVLA's 85.8\%: frequent contact leaves a substantial acquisition gap. Figure~\ref{fig:task-gap}b shows why reaching the target is insufficient: contact can end in escape or unstable holding.

\paragraph{Contact-physics sensitivity.}
We freeze checkpoints and repeat the same 600 starts per seed while changing one physical factor at a time. Relative to friction $\mu=0.6$ and restitution $e=0.05$, we test $\mu=0.3/0.9$, $e=0/0.15$, and joint mass--inertia scaling by $0.5/1.5$ at fixed geometry. With other settings fixed, Astro 0 retains 84.3--88.5\%, exceeding Direct RLT + ECC by 2.33--3.11 points (Fig.~\ref{fig:resource-controls}a). Across all three models and seven settings, contact exceeds retention by 8.9--17.6 points. The acquisition gap and ordering persist across these contact settings. Stable acquisition remains the harder endpoint throughout the sweep, rather than only under nominal contact parameters.

\subsection{From collection coverage to bias correction}
\label{sec:data-eval}
At the same 4,000-demonstration budget, we compare drift-only data with balanced coverage of all four nominal tracks (Table~\ref{tab:coverage-main}). Both use supervised adaptation only; training budgets and execution interfaces match within each model family. Balanced coverage improves retention across all three families. DynamicVLA sacrifices 2.5 points on drift but gains 9.8 on tumble and 15.2 on fast tumble. This reveals a specialization--coverage tradeoff across model families. Broader coverage helps tumbling at a small cost to drift specialization.

\begin{table}[!ht]
\centering\small
\caption{Equal-size motion coverage. Each model uses 4,000 demonstrations and supervised adaptation only. Retention (\%) pools 1,800 standard dynamic trials; gain is in percentage points.}
\label{tab:coverage-main}
\begin{tabular}{@{}lrrr@{}}
\toprule
Model & Drift only & Balanced & Gain\\\midrule
$\pi0.5$ & 66.8 & 74.5 & +7.7\\
SmolVLA & 68.8 & 76.4 & +7.6\\
DynamicVLA & 70.8 & 78.3 & +7.5\\
\bottomrule
\end{tabular}
\end{table}

\begin{figure*}[t]
\centering\includegraphics[width=\textwidth]{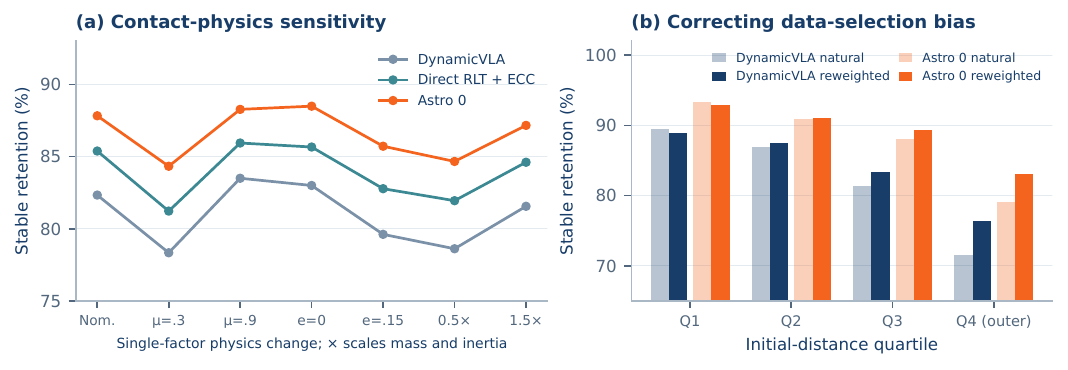}
\caption{\textbf{Contact sensitivity and sampling bias.} (a) Frozen models under one-factor physics changes. (b) Distance-stratified sampling with fixed demonstrations and budgets. Each setting uses three seeds and 1,800 evaluations; nominal controls reuse Table~\ref{tab:capture-overview}.}
\label{fig:resource-controls}
\end{figure*}

\begin{figure*}[t]
\centering\includegraphics[width=\textwidth]{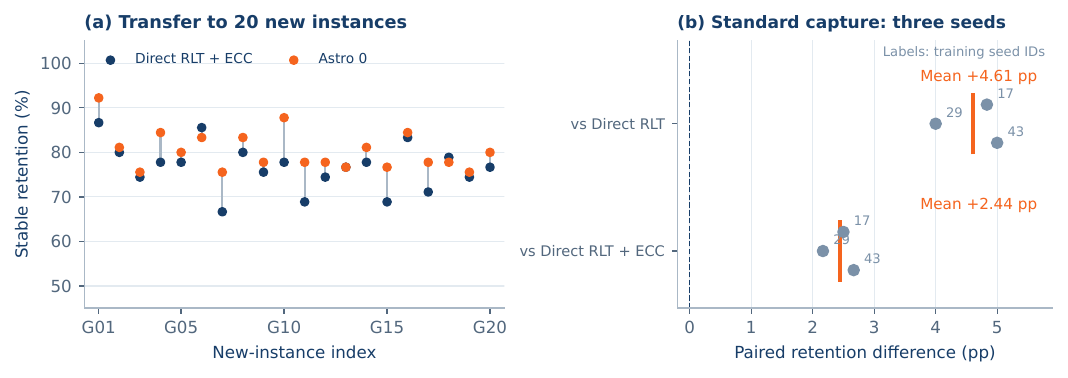}
\caption{\textbf{Transfer and paired gains.} (a) Retention on 20 new geometries. (b) Standard-retention gains for seeds 17/29/43; orange lines mark means. Panel (b) reuses Table~\ref{tab:capture-overview}.}
\label{fig:validation-v50}
\end{figure*}

\paragraph{Correcting success-selection bias.}
The outer distance quartile supplies 25\% of attempts but 19\% of accepted demonstrations. With all 16,000 trajectories and training budgets fixed, we sample distance quartiles uniformly within each object--motion cell, then sample an existing trajectory. Only demonstration stages change; Astro 0's online resets and interaction budget remain fixed.

Distance reweighting raises outer-quartile retention from 71.6 to 76.4\% for DynamicVLA and 79.1 to 83.1\% for Astro 0 (Fig.~\ref{fig:resource-controls}b). Overall rates rise from 82.3 to 84.1\% and 87.8 to 89.1\%; paired gains are 1.72 [0.68,2.76] and 1.28 [0.42,2.14] points. Slight inner-quartile declines reveal the allocation tradeoff. The unchanged trajectory pool shows that sampling existing demonstrations more evenly can improve capture without additional collection.

\paragraph{Beyond the fixed collector.}
The collector retains 360/600 standard states; extending only its lookahead from 2 to 4 s raises this to 398 and reduces planning refusals from 78 to 34. Collector success and failure define subsets of 1,080 and 720 executions across the three seeds. On the failure subset, Astro 0 retains 548 targets, compared with 487 for Direct RLT. This accounts for 61 of its 83 additional acquisitions overall. Finite search and closure gating limit the privileged collector. The supervised DynamicVLA reference already retains 82.3\%. We evaluate the further contribution of online learning with the same actor before and after that stage.

\subsection{Same-objective model comparison and transfer}
\label{sec:refinement-eval}
\begin{table}[!ht]\centering\small
\caption{BRO component ablation. The stochastic/uniform row is Direct RLT + ECC; the mean/balanced row is Astro 0. All variants use ECC, final-retention rewards, and equal tuning budgets. Rates and 95\% seed-$t$ intervals use three seeds and 1,800 trials/configuration.}
\label{tab:context-objective-main}
\setlength{\tabcolsep}{3.5pt}
\begin{tabular}{lrr}\toprule
Update / penalty & Retention & FCR\\\midrule
Stochastic / Uniform & 85.4 [82.9,87.9] & 79.0 [75.7,82.3]\\
Mean / Uniform & 86.4 [84.5,88.3] & 80.6 [78.3,82.8]\\
Stochastic / Balanced & 86.8 [84.7,88.9] & 81.6 [79.1,84.1]\\
Mean / Balanced & 87.8 [85.9,89.7] & 83.2 [80.7,85.7]\\
\bottomrule\end{tabular}
\end{table}

\begin{table}[!ht]\centering\small
\caption{Transfer under final-retention training. Brackets are 95\% seed-$t$ intervals from three training seeds. Combinations retrain on 4,000 demonstrations (1,200 trials/model); new geometry uses frozen 16,000-demo models (1,800 trials/model).}
\label{tab:v50-transfer}
\setlength{\tabcolsep}{3pt}
\resizebox{\columnwidth}{!}{\begin{tabular}{lrr}\toprule
Model & Retention & Target loss\\\midrule
\multicolumn{3}{l}{\textit{Unseen motion combinations}}\\
DynamicVLA ref. & 77.7 [74.2,81.1] & 16.0 [13.3,18.7]\\
Direct RLT & 81.0 [78.3,83.7] & 14.5 [11.8,17.2]\\
Direct RLT + ECC & 83.1 [79.9,86.3] & 12.5 [9.7,15.3]\\
Astro 0 & 86.0 [82.5,89.5] & 10.2 [7.0,13.5]\\
\midrule\multicolumn{3}{l}{\textit{New geometric instances}}\\
DynamicVLA ref. & 69.6 [66.5,72.7] & 21.7 [19.4,24.0]\\
Direct RLT & 74.5 [71.6,77.4] & 16.8 [15.3,18.3]\\
Direct RLT + ECC & 76.7 [74.0,79.4] & 15.3 [13.4,17.2]\\
Astro 0 & 80.3 [77.3,83.3] & 13.3 [11.7,15.0]\\
\bottomrule\end{tabular}}
\end{table}

\begin{figure*}[t]
\centering\includegraphics[width=\textwidth]{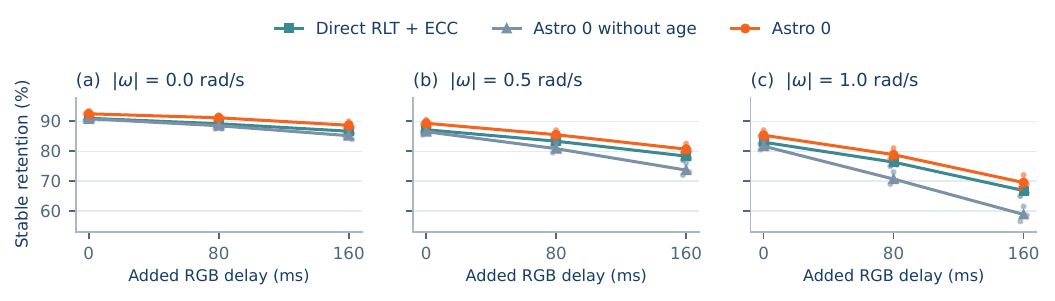}
\caption{\textbf{Visual delay and rotation.} Retention at 0.05 m/s translation; 200 poses and three seeds per setting. Lines show means, faint points seeds. Zero added delay retains native pipeline latency.}
\label{fig:visual-delay}
\end{figure*}
\paragraph{Execution context.}
ECC raises Direct RLT's retention from 1,498 to 1,537/1,800 (+2.17 points); BRO raises it to 1,581 (+2.44). Together they yield the full 4.61-point gain under matched rewards and tuning budgets. Thus execution context and the actor objective both contribute under the same final-retention reward.

\paragraph{BRO component ablation.}
Table~\ref{tab:context-objective-main} crosses actor updates and reference regularizers with ECC and reward fixed. BRO gains 44 retentions over stochastic/uniform updates; balancing alone adds 26 at fixed mean updates. Paired seed gains span 2.17--2.67 points over Direct RLT + ECC and 4.00--5.00 over Direct RLT (Fig.~\ref{fig:validation-v50}b). Each component helps when the other is fixed, and their combination performs best.

\paragraph{What changes after contact.}
Relative to Direct RLT + ECC, Astro 0 reduces unstable-contact failures from 97 to 60 and post-contact escapes from 119 to 117. Together with five fewer no-contact failures, these account for its 44 additional retentions. Both online models reduce escapes relative to DynamicVLA's 182. The additional BRO gain is concentrated in stabilizing contact, with little change in escape frequency.

\paragraph{Learning beyond imitation.}
Before online learning, the behavior-cloned actor retains 1,503/1,800 targets (83.5\%). Continuing training for 120k control steps raises retention to 87.8\%, a paired three-seed gain of 4.33 points [3.51,5.16], with the same frozen Token encoder and execution interface. Target loss falls from 207 to 138. Checkpoint and reference-strength selection use validation data only. Online refinement therefore contributes beyond imitation even when the visual representation remains frozen.

\paragraph{Visual age under rotation.}
We cross 0/80/160-ms added RGB delay with 0/0.5/1.0-rad/s rotation on 200 initial poses, matching capture cadence and execution clocks. Models train at nominal latency; the no-age variant zeros only the age input during training and evaluation. At 1 rad/s and zero added delay, retention is 81.7\% without age and 85.3\% with it. Under the same additional 160 ms, rates fall to 58.8\% and 69.5\% (Fig.~\ref{fig:visual-delay}). Thus the 10.67-point benefit compares identical delay conditions and grows as observations become older. Age conditioning mitigates stale evidence, although both variants still benefit from fresher observations.

\paragraph{Withheld motions and new instances.}
Combination models train on four pairs from $\{0.02,0.05\}$ m/s $\times\{0,0.5,1.0\}$ rad/s, excluding $(0.02,1.0)$ and $(0.05,0.5)$ from every learning stage. They use 4,000 demonstrations and 400 validation trajectories, including 1,000 training and 400 validation trajectories outside the core dataset. Frozen standard models face 20 new instances with ten starts per dynamic track.

Astro 0 reaches 86.0\% retention on withheld motions and 80.3\% on new instances (Table~\ref{tab:v50-transfer}), exceeding both RLT controls. G06/G18 favor Direct RLT + ECC (Fig.~\ref{fig:validation-v50}a). Excluding the near-variant G02 changes Astro 0's gain over that control from 3.67 to 3.80 points (1,710 evaluations/model). These are new instances, not unseen semantic categories. The overall gains extend beyond trained instances and motion pairs, with geometry-specific exceptions.

\section{Discussion and Conclusion}
\label{sec:limitations}
AstroBench exposes contact--retention gaps, collection bias, and visual-delay costs. Distance-stratified sampling and execution-aware refinement improve retention. Evaluation uses a fixed-base arm, limited objects and motions, and simulated contacts; physical transfer and coupled spacecraft--arm dynamics remain future work.

\label{main-text-end}
\clearpage
\section*{LLM Usage Statement}
ChatGPT/Codex assisted with drafting, proposing evaluation protocols and analysis plans, and figure preparation. An earlier model schematic used ImageGen; the current architecture diagram is vector-authored. The authors are responsible for verifying all claims, references, implementation details, and evidence.
{\small
  \bibliographystyle{ieeenat_fullname}
  \bibliography{references}
}
\end{document}